\documentclass[]{fairmeta}
\usepackage{wrapfig,lipsum}
\DeclareMathOperator*{\argmax}{arg\,max}
\newcommand*\circled[1]{\tikz[baseline=(char.base)]{
            \node[shape=circle,draw,inner sep=0.4pt] (char) {#1};}}
\usepackage{pifont}
\usepackage{tikz}
\newcommand{\xmark}{\ding{55}}%
\title{Stream RAG: Instant and Accurate Spoken\\ Dialogue Systems with Streaming Tool Usage}

\author[1,2,*]{Siddhant Arora}
\author[1]{Haidar Khan}
\author[1]{Kai Sun}
\author[1]{Xin Luna Dong}
\author[1]{Sajal Choudhary}
\author[1]{Seungwhan Moon}
\author[1]{Xinyuan Zhang}
\author[1]{Adithya Sagar}
\author[1]{Surya Teja Appini}
\author[1]{Kaushik Patnaik}
\author[1]{Sanat Sharma}
\author[2]{Shinji Watanabe}
\author[1]{Anuj Kumar}
\author[1]{Ahmed Aly}
\author[1]{Yue Liu}
\author[1]{Florian Metze}
\author[1]{Zhaojiang Lin}
\affiliation[1]{Meta}
\affiliation[2]{Carnegie Mellon University}

\contribution[*]{Work done at Meta}

\abstract{End-to-end speech-in speech-out dialogue systems are emerging as a powerful alternative to traditional ASR–LLM–TTS pipelines, generating more natural, expressive responses with significantly lower latency. However, these systems remain prone to hallucinations due to limited factual grounding. While text-based dialogue systems address this challenge by integrating tools such as web search and knowledge graph APIs, we introduce the first approach to extend tool use directly into speech-in speech-out systems. A key challenge is that tool integration substantially increases response latency, disrupting conversational flow. To mitigate this, we propose Streaming Retrieval-Augmented Generation (\emph{Streaming RAG}), a novel framework that reduces user-perceived latency by predicting tool queries in parallel with user speech,  even before the user finishes speaking. Specifically, we develop a post-training pipeline that teaches the model when to \emph{issue tool calls} during ongoing speech and how to generate spoken summaries that fuse audio queries with retrieved text results, thereby improving both \emph{accuracy and responsiveness}. To evaluate our approach, we construct AudioCRAG, a benchmark created by converting queries from the publicly available CRAG dataset into speech form. Experimental results demonstrate that our \emph{streaming RAG} approach increases QA accuracy by up to 200\% relative (from 11.1\% to 34.2\% absolute) and further enhances user experience by reducing tool use latency by 20\%. Importantly, our \emph{streaming RAG} approach is modality-agnostic and can be applied equally to typed input, paving the way for more agentic, real-time AI assistants.}

\date{\today}
\correspondence{\email{siddhana@andrew.cmu.edu}, \email{zhaojiang@meta.com}}

\begin{document}

\maketitle

\section{Introduction}
\label{section:intro}

Spoken Dialogue Systems (SDS) are foundational to many everyday technologies, powering intelligent assistants such as Alexa and Siri, as well as interactive voice response systems in customer service. With the rapid expansion of SDS capabilities to mobile phones and wearable devices, the need for robust, scalable, and generalizable solutions has never been greater. Traditionally, SDS have relied on cascaded pipelines composed of multiple modules—including voice activity detection (VAD), automatic speech recognition (ASR), natural language understanding (NLU), natural language generation (NLG), and text-to-speech (TTS) synthesis—each introducing potential points of failure and latency \citep{glass1999challenges,huang2024audiogpt}.
Recently, end-to-end (E2E) SDS \citep{xie2024miniomnilanguagemodelshear,Dialog_GSLM,meng2024parrot,zhang2024turnbasedgameenablingrealtime,arora2025cotsds} have been proposed, which directly generate spoken responses from speech input within a unified architecture. This E2E approach not only mitigates error propagation across modules but also captures non-phonemic information more effectively, resulting in significantly lower inference time and computational overhead, and paving the way for more natural and efficient conversational experiences.

Despite these advances, current E2E SDS are fundamentally constrained by their reliance on internalized knowledge from static training data, which often results in responses that lack factual grounding or fail to reflect the most up to date information. This shortcoming is particularly critical for action-oriented or knowledge-seeking tasks, such as booking hotels or answering questions about current events. In contrast, text-based conversational assistants have begun to overcome these limitations by integrating external tools through Retrieval-Augmented Generation (RAG)\citep{yang2024crag,RAG1,RAG2,RAG3,RAG4}, dynamically retrieving relevant information from sources like web search, knowledge graphs (KG), and real-time APIs. Yet, the integration of such tool use into E2E SDS remains largely unexplored.
A key challenge is that while external tools can substantially improve factual accuracy, invoking them often introduces additional latency, leading to awkward silences that disrupt the natural conversational flow. This raises a research question: {\em How can we trade-off between accuracy and responsiveness for developing SDS that feel both intelligent and natural?}

\begin{wrapfigure}{r}{0.5\linewidth}
\vskip -0.25in
\begin{center}
\vskip -0.25in
\includegraphics[width=\linewidth]{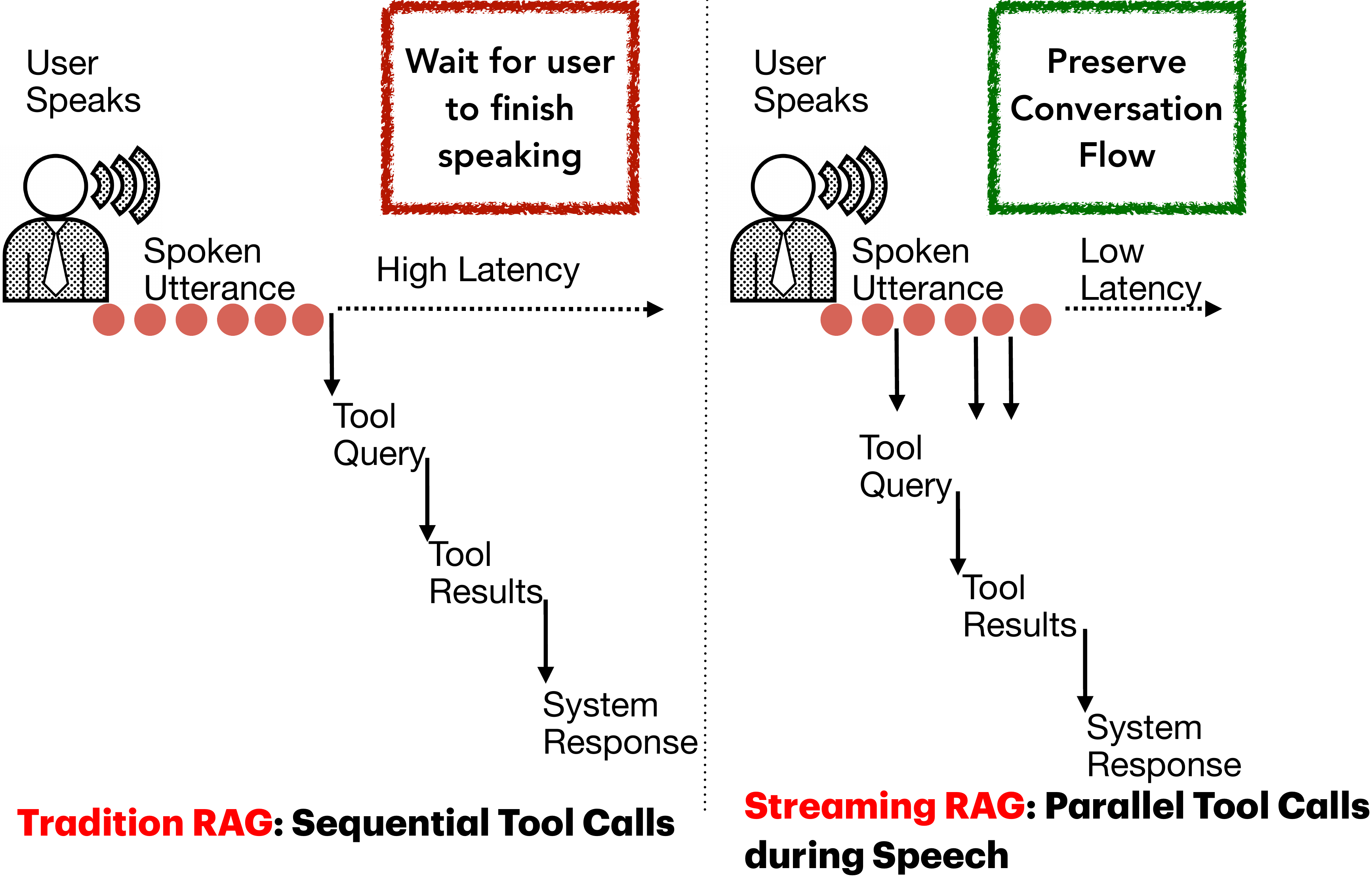}
\end{center}
\caption{Comparison of Traditional RAG with proposed Streaming RAG which fires tool queries in parallel with user speech.}
\label{fig:motivate_fig}
\vskip -0.18in
\end{wrapfigure}

In this paper, we present, to the best of our knowledge, the first {\em speech-in, speech-out language model that seamlessly integrates external tool invocation with low latency}. The key idea is a {\bf Streaming RAG} strategy that generates tool queries in parallel with user speech, often times even before the user has finished speaking (Fig.~\ref{fig:motivate_fig}). A naive implementation of streaming queries, however, faces two challenges: (1) queries issued from partial speech may be suboptimal, yielding distracting tool outputs and inaccurate responses; and (2) unnecessary tool calls may be triggered, wasting computational resources. We introduce effective modeling techniques to address these challenges and make the following contributions.

\textbf{Contribution 1}: We introduce a formal framework for tool integration in speech-in speech-out systems and empirically show that leveraging web search and KG APIs significantly enhances factual question answering. Evaluating three state-of-the-art (SOTA) models, Qwen-OMNI \citep{Qwen2.5-Omni}, OpusLM \citep{tian2025opuslm}, and Kimi-Audio \citep{ding2025kimi}, we find that external tool integration delivers substantial performance gains, boosting accuracy by up to 140\% relative (from 11.1\% to 26.3\% absolute). However, tool usage also introduces considerable latency, increasing first-token response time by 2.3x.

\textbf{Contribution 2}: To address this, we propose \emph{Streaming Retrieval-Augmented Generation} (\emph{Streaming RAG}), the \emph{first} framework that empowers the system to trigger tool queries in parallel with user speech, even before the user finishes speaking. Within this framework, we introduce two novel approaches: (1) \emph{Fixed-Interval Streaming RAG}, which issues tool queries at regular intervals during speech input and carefully examines quality of retrieval results on the full query to guarantee response quality, and can be incorporated into any speech-in, speech-out model without post-training; (2) \emph{Model-Triggered Streaming RAG}, which post-trains the model to intelligently determine optimal query timing based on the evolving user utterance to save computation resources. 
Our results demonstrate that our proposed \emph{Model-Triggered Streaming RAG} delivers over 200\% relative improvement in accuracy (from 11.1\% to 34.2\% absolute in T.~\ref{table:main_model_performance}) compared to the no-tool baseline, while also reducing tool result generation latency by 20\%. Though designed for speech-in speech-out systems, streaming RAG can also be adapted in cascaded SDS, or even chatbots as users type.

\textbf{Contribution 3}: Finally, we introduce AudioCRAG, a benchmark created by recording spoken queries from the CRAG \citep{yang2024crag} dataset, enabling robust evaluation of tool usage capabilities in speech-in speech-out systems. 
To support future research, we will open source our training code and AudioCRAG-Human benchmark, supporting future research in tool-integrated voice assistants.

\section{Related studies}
\label{sec: related works}

\subsection{Benchmarks for Tool Usage}
\label{subsec:related_dialogue_tool}
Recent advances in benchmarking text-based dialogue systems for tool usage \citep{text_rag_bench5,text_rag_bench1,text_rag_bench2,text_rag_bench3,text_rag_bench4} have primarily focused on evaluating factual question answering and task completion within simulated environments (detailed related work discussion in Appendix~\ref{subsec:related_text_dialogue_tool},~\ref{subsec:related_multimodal_tool}). The CRAG benchmark \citep{yang2024crag} is a leading example, featuring 4,409 question-answer pairs and providing mock APIs for both web and KG search.  Recent benchmarks \citep{cragmm2025,ma2024mms,videowebarena} have extended tool-augmented dialogue evaluation to multimodal input and longer-context scenarios. While these benchmarks have significantly advanced the evaluation of tool-augmented dialogue systems, they remain largely limited to text-based outputs and do not fully address the unique challenges presented by speech-in speech-out systems.

\subsection{E2E Spoken Dialogue Systems}
\label{subsec:spoken_dialogue_systems}
Several E2E spoken dialogue systems \citep{Qwen2.5-Omni,xie2024miniomnilanguagemodelshear,arora2025landscapespokenlanguagemodels,Dialog_GSLM,meng2024parrot,zhang2024turnbasedgameenablingrealtime,arora2025cotsds} have recently been introduced, demonstrating impressive semantic understanding and high audio quality in their responses. However, these systems have not yet been trained or evaluated for their ability to \emph{use external tools}.
Another research direction \citep{feng2025enhancing} explores E2E RAG for direct speech-to-text retrieval, utilizing multimodal embeddings for enabling speech utterances to directly retrieve relevant text. While this method outperforms models without RAG, it is primarily limited to \emph{speech-to-text} scenarios. Its ability to access KGs and other APIs, critical for real-world applications, remains unproven. Furthermore, the retrieval scope is limited, as experiments are conducted with retrieval restricted to just 10 paragraphs, whereas modern RAG benchmarks require searching across thousands of web pages.
Although recent studies \citep{maben2025aura} have developed web interfaces that integrate tools into speech-in \emph{speech-out} scenarios using cascaded pipelines, comprehensive empirical investigations of E2E speech-in speech-out systems and, crucially, systematic analyses of user-perceived latency, remain largely unexplored. In this work, we address these gaps by developing a comprehensive framework for tool integration in E2E speech-in speech-out systems and designing benchmarks to quantitatively assess the tool usage capabilities of SOTA models. Furthermore, recognizing the importance of latency for natural conversational flow, we introduce novel streaming RAG methods that not only enhance tool usage performance but also reduce user-perceived latency.

\section{Methodology}
A RAG spoken conversation system takes an audio question $Q$ as input and outputs a spoken answer $A$. Let the ASR transcript of audio question be $X^{\text{asr}}$ and the ASR transcript of the spoken answer be $X^{\text{res}}$.
Answers are generated by speech-in speech-out models, leveraging both the model’s internal knowledge and information retrieved from external sources.
To incorporate external information, the model needs to formulate a tool query $Q^{T}$ to retrieve relevant results $R$ from an external tool $T$. 

\subsection{Tool Integration for Speech-in Speech-Out LLMs}
\label{subsec:tool_integrate_method}
\label{sec:method}
\begin{wrapfigure}{r}{0.5\linewidth}
\vskip -0.25in
\begin{center}
\vskip -0.2in
\includegraphics[width=\linewidth]{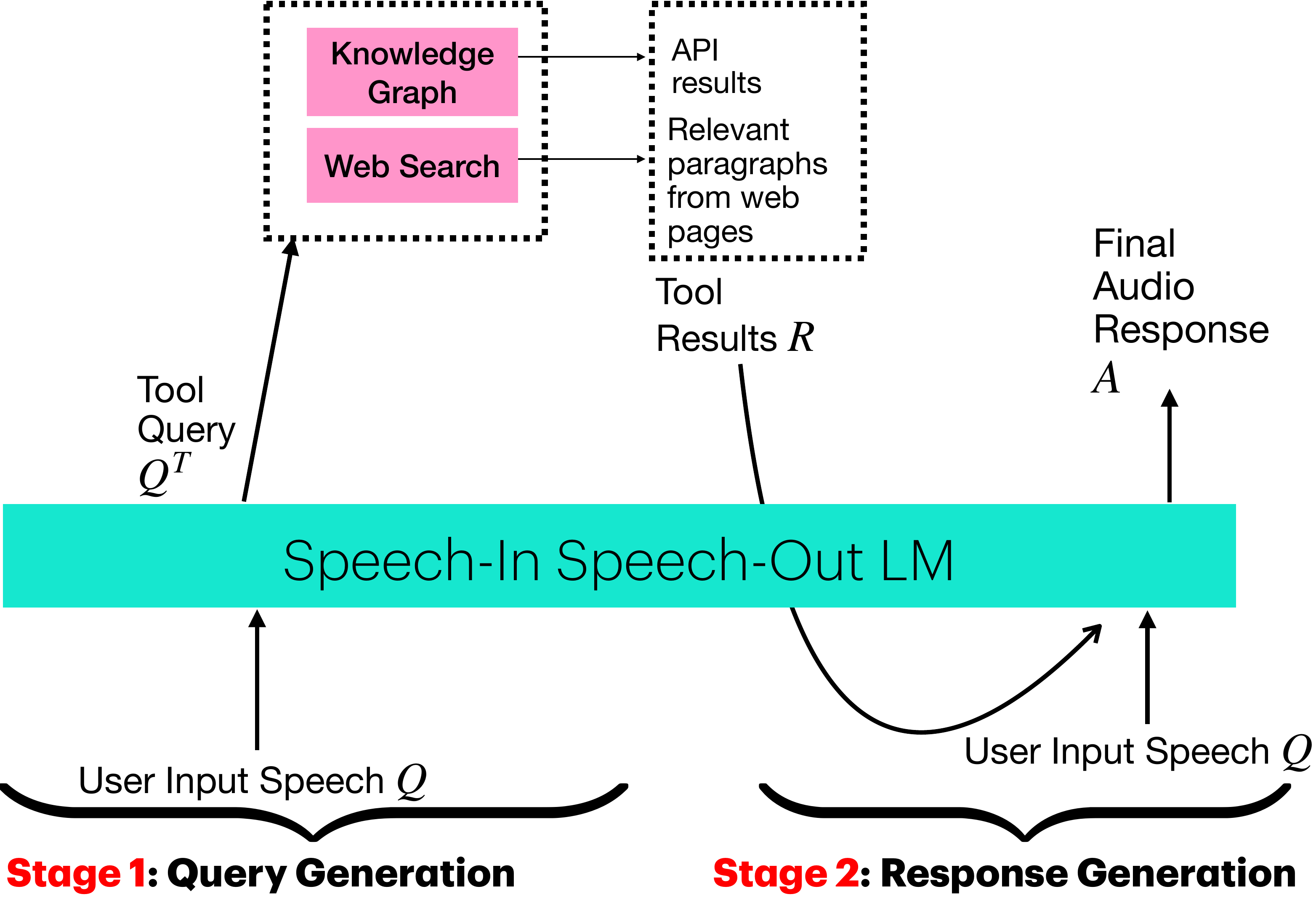}
\end{center}
\caption{Proposed formulation for integrating tool usage in E2E speech-in, speech-out dialogue systems using a two-stage inference approach.}
\label{fig:tool_integration_formulate}
\vskip -0.15in
\end{wrapfigure}
Figure~\ref{fig:tool_integration_formulate} illustrates our proposed formulation for integrating external tools into speech-in speech-out systems. We introduce a two-stage inference approach: \emph{Query Generation} and \emph{Response Generation}. In the Query Generation stage, the system processes an audio question and generates queries for each external tool to retrieve relevant information by maximizing the posterior distribution $P(Q^{T}|Q)$ (Examples of generated queries are provided in T.~\ref{tab:example_queries} in the Appendix.). In the Response Generation stage, the retrieved results $R$ from these tools are combined with the original audio question and input into the model to generate the final spoken response by maximizing the posterior distribution $P(A|Q,R)$. By conditioning the final output generation on the input audio, this formulation not only provides a simple and effective mechanism for interacting with text-based APIs, but also preserves the key advantages of speech-in speech-out systems, mitigating error propagation and enabling the model to capture non-phonemic information (such as prosody and speaker intent) more effectively.
\subsection{Streaming RAG}
\label{sec:stream_rag}
\begin{figure}[t]
    \centering
    \begin{subfigure}[b]{0.48\linewidth}
        \centering
        \includegraphics[width=\linewidth]{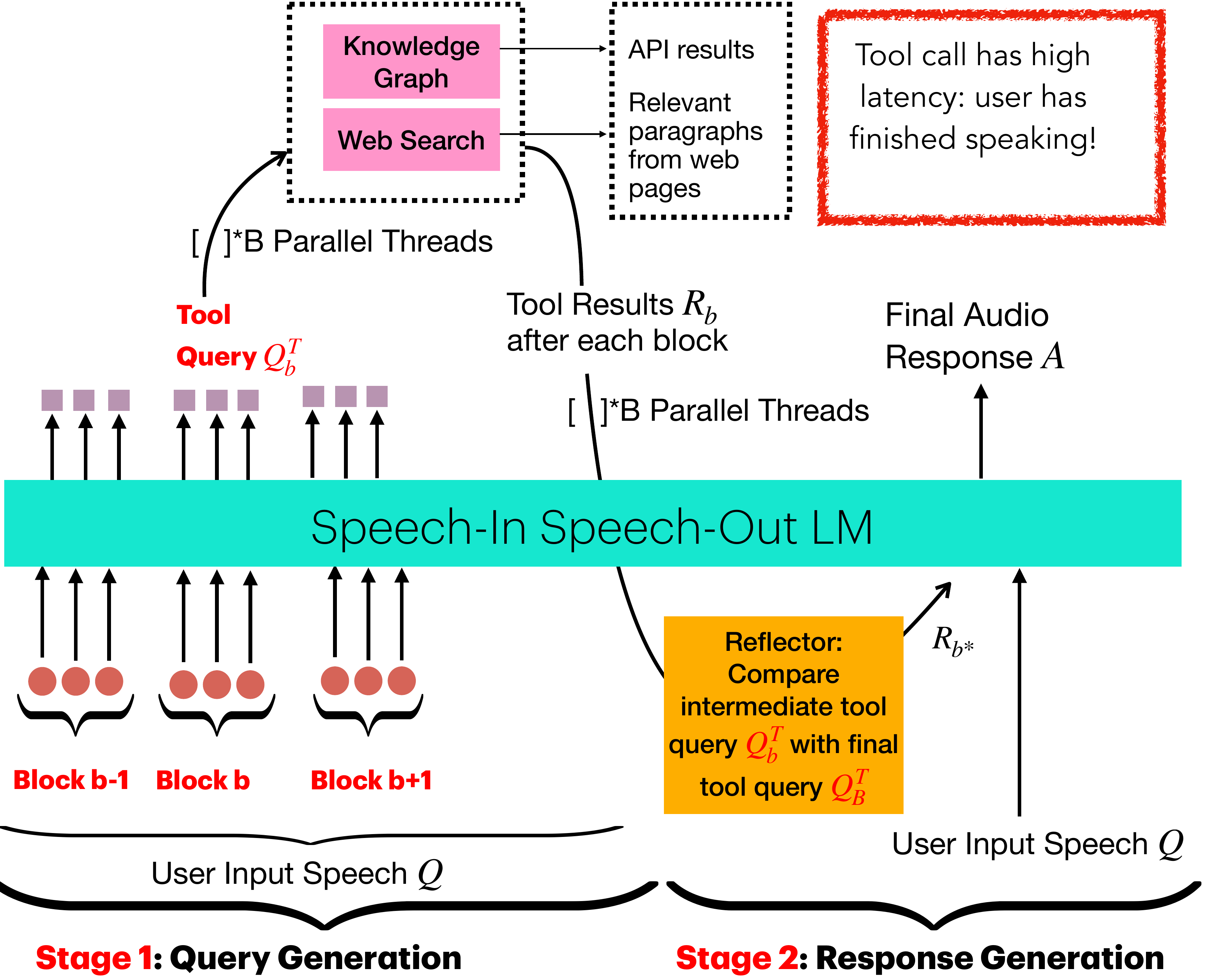}
        \caption{Fixed-Interval Streaming RAG.}
        \label{fig:streaming_rag_inference}
    \end{subfigure}
    \hfill
    \begin{subfigure}[b]{0.48\linewidth}
        \centering
        \includegraphics[width=\linewidth]{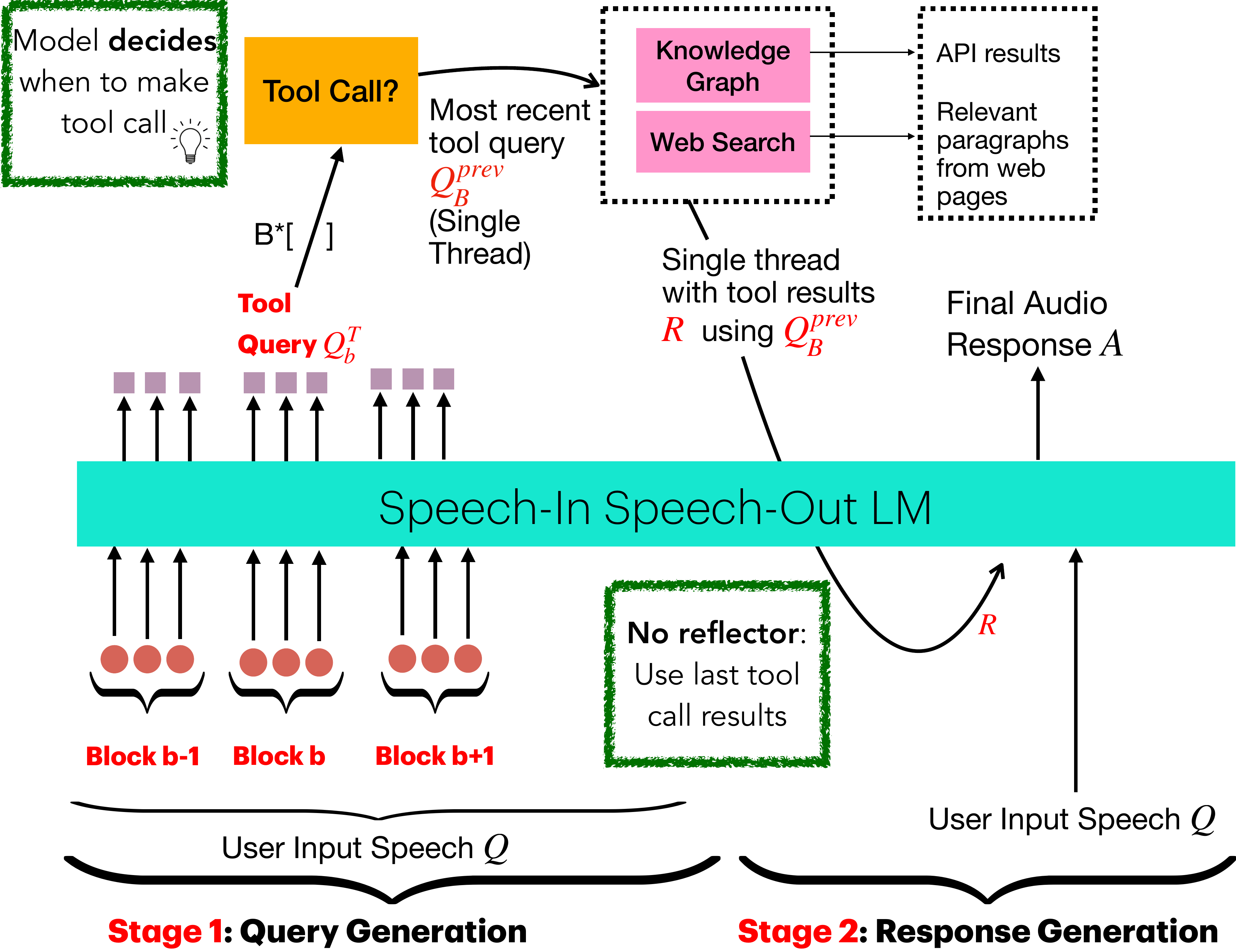}
        \caption{Model-Triggered Streaming RAG.}
        \label{fig:post_train_streaming_rag}
    \end{subfigure}
    \caption{Proposed formulations for streaming tool query generation, referred to as \emph{Streaming RAG}, to minimize user-perceived latency in speech-in, speech-out systems. (a) Fixed-Interval Streaming RAG: tool calls are triggered at fixed intervals and evaluated by a reflector module. (b) Model-Triggered Streaming RAG: the model autonomously decides when to make tool calls, eliminating the need for a reflector and directly utilizing the most recent tool results for response generation.}
    \label{fig:streaming_rag_formulate}
\end{figure}

RAG-based systems, as proposed in S.~\ref{subsec:tool_integrate_method}, can significantly improve factual accuracy by incorporating external tools. However, these tool calls often introduce substantial latency, which is particularly problematic in speech-in, speech-out applications where users expect rapid, conversational responses and even brief silences can disrupt the natural flow of dialogue. 
One way to address this challenge lies in the nature of audio inputs, which arrive as a continuous stream. This streaming property enables tool calls to be initiated before the user has finished speaking, offering a unique opportunity to mitigate latency.

To minimize user-perceived latency, we introduce \emph{Streaming RAG}: the first framework to generate and issue tool queries in parallel as audio input is received. This novel approach is built on three key design components:
\circled{1} Trigger: When to initiate a new tool query;
\circled{2} Threads: The number of parallel tool query threads;
\circled{3} Reflector: The module that determines whether intermediate tool results are sufficient for generating the final output. By exploring different design choices for each component, we introduce two complementary approaches for streaming tool query generation in the following subsections: a fixed-interval trigger method and a model-based trigger method. 

\subsubsection{Fixed-Interval Streaming RAG}
\label{subsec:stream_rag_infer_method}
In this approach, the \emph{trigger} is set to fire tool calls at fixed chunk intervals during audio input.
The input speech $Q$ is divided into a sequence of $B$ blocks, $Q = \{Q_b \mid b=1, \dots, B\}$, with each block containing $N_{\text{block}}$ frames.
To approximate $P(Q^T|Q)$ as described in S.~3.1, we follow a block-wise prediction strategy. In this approach, after processing each audio block $b$, the model predicts a tool query $\hat{Q}_b^T$ by conditioning on the input speech accumulated up to block b, specifically $Q_{1:b}$:
\begin{equation}
\hat{Q}_b^T  = \argmax_{Q_b^T} P(Q_b^T|Q_{1:b}) \label{eq:infer_stream_RAG}
\end{equation}
This strategy results in $B$ parallel tool call threads running simultaneously (see Figure~\ref{fig:streaming_rag_inference}), where each thread generates a tool query prediction $\hat{Q}_b^T$ for its corresponding block $b \in [1,B]$.
The tool queries $\hat{Q}^{T}_{b}$ generated after each block are then stored in cache. Given the high latency of tool calls (See T.~\ref{tab:latency}), users typically complete their utterances before tool responses are ready. Upon utterance completion, an explicit \emph{reflector} module ``$\text{reflect}()$'' evaluates the cached intermediate queries $\hat{Q}^{T}_{b}$ against final query $\hat{Q}^{T}_{B}$ to determine whether an early intermediate tool call provides sufficient information to answer the user's question $Q$. The reflector module systematically evaluates all intermediate queries in the cache and identifies the \emph{earliest} sufficient tool call $b^\star$ where the intermediate query $\hat{Q}^{T}_{b}$ will give the same result as final query $\hat{Q}^{T}_{B}$ as shown:
\begin{equation}
b^\star = \min \{b \in [1,B] \text{ where reflect}(\hat{Q}^{T}_{b},\hat{Q}^{T}_{B}) = \mathrm{True}\}.
\label{eq:reflector_select}
\end{equation}
All subsequent parallel tool calls after $b^\star$ are promptly terminated, and the retrieved results $R_{b^\star}$ from this intermediate call $\hat{Q}^{T}_{b^\star}$ are used to generate the final spoken response $A$ by maximizing the posterior distribution $P(A|Q,R_{b^\star})$ (instead of $P(A|Q,R)$ in S.~\ref{subsec:tool_integrate_method}).
We employ a reflector module that uses simple yet effective heuristics: (a) if the top 5 web documents for an intermediate web query match those of the final web query, and (b) if the KG results for intermediate and final KG queries are identical. 
These heuristics ensure that the information retrieved from an early tool call using $\hat{Q}^{T}_{b^\star}$ is consistent with what would have been obtained by waiting for the final tool call using $\hat{Q}^{T}_{B}$, thereby providing a strong quality guarantee. Since most tool call latency arises from chunking and reranking the chunks of web documents, these checks enable significant latency savings without any compromise in model performance.
A key advantage of this strategy is its plug-and-play nature: it requires no additional post-training for speech-in, speech-out models and can be directly applied at inference time across a variety of architectures. However, there are important considerations. First, generating parallel tool calls at every fixed interval (Eq.~\ref{eq:infer_stream_RAG}) increases computational overhead, which may pose challenges for deployment on resource-constrained devices such as wearables devices. Second, reliance on an external reflector module (Eq.~\ref{eq:reflector_select}) to determine the sufficiency of intermediate tool calls may limit the extent of achievable latency improvements. 

\subsubsection{Model-Triggered Streaming RAG} 
\label{subsec:streaming_rag_post_training_method}
To address the limitations of Fixed-Interval Streaming RAG and further optimize both efficiency and responsiveness, we propose a more adaptive approach: \emph{Model-Triggered Streaming RAG}.
Here, the \emph{trigger} is learned: the model is trained to autonomously determine the optimal moments to initiate tool queries, issuing a query only when it encounters new or additional information as illustrated in Figure~\ref{fig:post_train_streaming_rag}. In this formulation, the model receives user input in fixed chunk intervals as before and intelligently determines whether a tool call is needed after each block $b$. The model can either: \circled{1} Predict 
NO\_QUERY if a new tool query is unnecessary, or
\circled{2} Generate a new tool query. To make this decision, the model conditions on both the accumulated input speech $Q_{1:b}$ (see Eq.~\ref{eq:infer_stream_RAG}) as well as most recent tool query $\hat{Q}^{\text{prev}}_{b}=\hat{Q}^{T}_{\max\{\, i < b : \hat{Q}^{T}_i \neq \text{NO\_QUERY} \,\}}$ as shown:
\begin{equation}
\begin{aligned}
\hat{Q}^T_b =\argmax_{Q^T_b} P(Q^T_b|Q_{1:b},\hat{Q}^{\text{prev}}_b)
\end{aligned}
\label{eq:stream_RAG_post_train}
\end{equation}
When a new query $\hat{Q}^{T}_{b} \neq \text{NO\_QUERY}$, the system immediately terminates any ongoing tool calls for the previous query  $\hat{Q}^{\text{prev}}_{b}$, ensuring that only a \emph{single tool call thread} runs at any given time. (Examples of generated $\hat{Q}^{T}_{b}$ are provided in T.~\ref{tab:example_stream_queries} in the Appendix.)
This approach offers several key advantages. First, it effectively eliminating redundant parallel threads and significantly reducing computational overhead. This is especially important for deployment on resource-constrained devices.
Second, this formulation removes the need for an external \emph{reflector} (Eq.~\ref{eq:reflector_select}) module. The model confidently relies on the results $R$ from the most recent tool call using $\hat{Q}^{\text{prev}}_{B}$ to generate the spoken response $A$ by maximizing $P(A|Q,R)$, reducing system complexity.

\textbf{Post-training}: To train the model, we transform text-based tool usage benchmarks into spoken format (See S.~\ref{subsec:post_train_data_prep}). Word-level timestamps are computed using a pre-trained ASR model. For each partial ASR transcript $X^{\text{asr}}_b$ up to block $b$,  we generate corresponding queries  for each tool $\overline{Q^{T}_{b}}$ using an LLM as pseudo ground truth (GT). 
To create effective training labels that teach the model when to trigger new queries, we employ a similarity-based labeling strategy where we compare the current pseudo GT query $\overline{Q^{T}_{b}}$ with the most recent non-empty tool query label $\hat{Q}^{\text{prev}}_b$ (see Eq.~\ref{eq:stream_RAG_post_train}) before block $b$. 
Our labeling function assigns the training label $\hat{Q}^{T}_{b}$ (Eq.~\ref{eq:stream_RAG_post_train}) for the tool query after block $b$ as follows: \circled{1} when the current query is sufficiently similar to the previous query (as determined by manually defined heuristics  $f(\cdots)$), we assign the special label NO\_QUERY to teach the model that no new tool call is needed. \circled{2} When the queries are sufficiently different, we assign the actual pseudo ground truth query 
$\overline{Q^{T}_{b}}$  as the label to teach the model to trigger a new tool call:
\begin{equation}
    \hat{Q}^{T}_{b} =
\begin{cases}
\text{NO\_QUERY}, & \text{if } f(\overline{Q^{T}_{b}}, \hat{Q}^{\text{prev}}_{b})= \mathrm{True}, \\[2pt]
\overline{Q^{T}_{b}}, & \text{else }.
\end{cases}
\label{eq:stream_RAG_label}
\end{equation}
For KG queries, we assign a NO\_QUERY label when the current query exactly matches the previous one. For web queries, we assign a NO\_QUERY label if the top five retrieved documents for the current query remain unchanged from the previous query.  

We employ a multi-task fine-tuning strategy targeting two key capabilities.
First, we train the model on Streaming Tool Query Generation by optimizing $P(Q^T_b|Q_{1:b},\hat{Q}^{\text{prev}}_b)$ for $b \in [1,B]$, enabling intelligent decisions about when to trigger tool queries. Second, we fine-tune on Response Generation by optimizing $P(A|Q,R)$ (S.~\ref{subsec:tool_integrate_method}) to improve the intelligibility of the speech output.

An important aspect of our post-training is enhancing the model’s ability to recover from errors in intermediate query predictions. For example, when presented with the audio question, ``Who founded Rare Beauty in 2019?'', we observed that an initial misinterpretation of $\hat{Q}^{\text{prev}}_{b}$, such as ``Red Bull founder'', can lead the model to subsequently generate NO\_QUERY labels, effectively halting further attempts to retrieve the correct information. This issue arises because, during training, the model is always provided with correct previous query $\hat{Q}^{\text{prev}}_{b}$, whereas during inference, it may make errors due to partial utterances ${Q}_{1:b}$ being ambiguous. Thus the model lacks the ability to recover from such mistakes during inference.
To overcome this, we introduce a novel strategy in which we deliberately inject negative samples during post-training by substituting the previous query $\hat{Q}^{\text{prev}}_{b}$ in Eq.~\ref{eq:stream_RAG_label} with incorrect ones $Q^{\text{neg}}_{b}$. Crucially, when we perform negative sampling, we fall back to the pseudo ground truth query $\overline{Q^{T}_{b}}$ as the training label:
\begin{equation}
(\hat{Q}^{T}_{b},\hat{Q}^{\text{prev}}_{b})=
\begin{cases}
 (\hat{Q}^{T}_{b}, \hat{Q}^{\text{prev}}_{b}), & \text{with probability } 0.9, \\[2pt]
(\overline{Q^{T}_{b}}, Q^{\text{neg}}_{b}), & \text{with probability } 0.1.
\end{cases}
\label{eq:stream_RAG_neg_sample}
\end{equation}
This approach explicitly teaches the model to recover from errors in intermediate query prediction, thereby maintaining accuracy (see T.~\ref{tab:streaming_rag_ablatiob_results} for ablation) while achieving latency savings.
\begin{table}[t]
\centering
\caption{Performance comparison of accuracy, first-token latency, and latency savings (as a percentage of tool use latency, which is 3.37 seconds as reported in Table~\ref{tab:latency}) for three models—Qwen2.5-7B, OpusLM, and Kimi Audio—across three settings: Closed Book (without tool usage), Open Book (with tool usage), and Streaming RAG (\emph{Model-Triggered Streaming RAG}). We evaluate all models on both the AudioCRAG-Synthetic (Syn.) and AudioCRAG-Human (Hum.). Streaming RAG is not applied to Kimi-Audio as it can handle only a restricted length of tool result references (S.~\ref{sec:app-1}). *: OpusLM currently does not support taking tool result references in speech-out settings in zero-shot.}
\begin{tabular}{lllllllll}
\toprule
Setting &Ref length & Model & \multicolumn{2}{c}{Accuracy} & \multicolumn{4}{c}{Latency}  \\
& & & Syn. & Hum. & \multicolumn{2}{c}{First Token (s)} & \multicolumn{2}{c}{\% Savings} \\
& & & & & Syn. & Hum. &  Syn. & Hum. \\
\midrule
Closed Book & 0          & Qwen2.5-7B & 11.1 & 13.1 & 1.34 & \hphantom{0}1.24 & \xmark &\xmark \\
& 0     & OpusLM     & 18.4 & 15.5 &  5.67  & \hphantom{0}7.07 & \xmark &\xmark\\
& 0  & Kimi Audio &  16.7 & 16.0 & 0.85  & \hphantom{0}0.89 &\xmark  &\xmark \\
\midrule
Open Book & 23K & Qwen2.5-7B & 26.3 & 26.9 & 5.90 & \hphantom{0}5.40 &\xmark &\xmark\\
(S.~\ref{subsec:tool_integrate_method}) & 15K & OpusLM* & \hphantom{0}0.0 & \hphantom{0}0.0 & 9.05 & 10.44 &\xmark &\xmark\\
& 500 & Kimi Audio & 21.8 & 19.6 & 4.22 & \hphantom{0}4.22 &\xmark &\xmark\\
\midrule
Streaming RAG & 23K & Qwen2.5-7B & 34.2 & 37.4 & 5.32 & \hphantom{0}3.60 & 20.7\% & 53.4\%\\
(S.~\ref{subsec:streaming_rag_post_training_method}) & 15K & OpusLM & 23.6 & 22.8 &  8.63 & \hphantom{0}9.04  &14.8\% & 41.5\%\\
\bottomrule
\end{tabular}
\label{table:main_model_performance}
\end{table}

\section{Experiment Setup}
\label{sec:experiment_setup}

\subsection{Evaluation Benchmarks}
To rigorously evaluate our proposed approach, we construct comprehensive benchmark datasets featuring spoken queries paired with simulated tool interactions. We begin with the CRAG dataset \citep{yang2024crag}, which form the basis for our spoken version of CRAG, which we term \emph{AudioCRAG}. It consists of 2 distinct variants: \circled{1} \textbf{Audio CRAG Synthetic}:  To generate spoken queries, we use our in-house TTS system. We then apply a rigorous filtering procedure which results in a high-quality set of 1,862 spoken queries, which we refer to as the \emph{AudioCRAG-Synthetic} benchmark. \circled{2} \textbf{Audio CRAG Human}: To further enhance the realism and diversity of our evaluation, we introduce the AudioCRAG-Human benchmark, which consists of 618 human-recorded spoken queries. We will release
Audio CRAG Human benchmark upon acceptance to supports the development of more natural and reliable voice assistants.
Further details on the construction of these benchmarks are provided in Sec.~\ref{subsec:audio_crag_appendix}. We follow the CRAG setup to incorporate both web and KG-based tools, and adopt its robust evaluation methodology, as described in Secs.~\ref{subsec:tool_usage_setup} and~\ref{subsec:evaluation_setting}.
Additionally, we leverage a random subset of 16,000 questions from the text-based factual question answering dataset TriviaQA~\citep{triviaqa} to post-train our speech-in, speech-out models, as detailed in Sec.~\ref{subsec:post_train_data_prep}.

\subsection{Evaluated SOTA Speech-in Speech-out Models}
\label{subsec:baseline_models}
In this work, we present a comprehensive benchmark of three SOTA speech-in, speech-out conversational systems: Qwen-OMNI~\citep{Qwen2.5-Omni}, Kimi-Audio~\citep{ding2025kimi} and OpusLM~\citep{tian2025opuslm}.
We evaluate them under both tool-augmented and non-tool-augmented conditions. Further details about our experimental setup are provided in S.~\ref{sec:app-1}.

We perform an ablation study (referred to as ``Tool Integration'' in T.~\ref{tab:post_train_ablation_performance}) where we post-train the model on sequential query generation (i.e. $P(Q^{T}|Q)$ in S.~\ref{subsec:tool_integrate_method}) and output generation, to assess the impact of streaming RAG post-training versus standard post-training on final response generation.
Additionally, we conduct an ablation study on open book setting (S.~\ref{subsec:tool_integrate_method}) using a self-cascade approach with a three-stage inference pipeline: (1) the audio question is used to generate a tool query $Q^{T}$ (corresponding to the ``Query Generation'' stage described in S.~\ref{subsec:tool_integrate_method}); (2) the audio question, and tool results are combined to produce the final text output $X^{\text{res}}$ by maximizing $P(X^{\text{res}}|Q,R)$; and (3) the audio question, and final text output are used to generate the final speech output $A$ by optimizing $P(A|Q,X^{\text{res}})$. Since we teacher-force the text output to obtain the final speech output in stage (3), this self-cascade approach can only be applied to a ``thinker-talker'' architecture (eq. Qwen-OMNI) or Chain-of-Thought (CoT) style architectures (eg. OpusLM).
The motivation for this ablation is to investigate whether the inclusion of RAG references $R$ during the ``Response Generation'' stage (S.~\ref{subsec:tool_integrate_method}) affects the quality of the generated speech $A$.

\begin{table}[t]
\centering
\caption{Results on AudioCRAG-Synthetic for Qwen2.5-7B, OpusLM, and Kimi Audio comparing text vs. speech output across Closed Book, Open Book, and \emph{Model-Triggered Streaming RAG}.}
\begin{tabular}{lllll}
\toprule
Setting & Ref length & Output & Model       & Acc.  \\
\midrule
Closed Book & 0          & Text   & Qwen2.5-7B & 15.0   \\
 & 0          & Speech & Qwen2.5-7B & 11.1  \\
 & 0          & Text   & OpusLM     & 20.1   \\
 & 0          & Speech & OpusLM     & 18.4 \\
 & 0          & Text   & Kimi Audio & 24.2  \\
 & 0          & Speech & Kimi Audio & 16.7 \\\midrule
Open Book   & 23K & Text & Qwen2.5-7B & 39.6  \\
 (S.~\ref{subsec:tool_integrate_method})  & 23K & Speech & Qwen2.5-7B & 26.3 \\
 & 23K & Speech (self-cascade) & Qwen2.5-7B & 33.8\\
 & 15K & Text & OpusLM & 26.3 \\
 & 15K & Speech & OpusLM & \hphantom{0}0.0  \\
 & 15K & Speech (self-cascade) & OpusLM & 21.2 \\
 & 5K & Text & Kimi Audio & 45.8 \\
 & 500 & Speech & Kimi Audio & 21.8 \\ \midrule
Streaming RAG  & 23K & Text & Qwen2.5-7B & 39.8  \\
(S.~\ref{subsec:streaming_rag_post_training_method})  & 23K & Speech & Qwen2.5-7B & 34.2\\
 & 15K & Text & OpusLM &  29.7 \\
 & 15K & Speech & OpusLM & 23.6\\
\bottomrule
\end{tabular}
\label{table:speech_to_text_performance}
\end{table}

\section{Results}
\subsection{Impact of Tool Integration and Streaming RAG on SOTA Models}
\label{subsec:results_rag}
Table~\ref{table:main_model_performance} provides a comprehensive performance comparison of three models, Qwen2.5-7B, OpusLM, and Kimi Audio, evaluated across three settings: Closed Book, Open Book, and Streaming RAG. All models are assessed on both the AudioCRAG-Synthetic (Syn.) and AudioCRAG-Human (Hum.). In the Closed Book setting, where models rely solely on their internal knowledge without access to external tools (reference length = 0), all models achieve accuracy scores below 20\%. These results highlight the inherent limitations of closed-book approaches in handling complex queries.
The Open Book setting, which provides models with access to external information, demonstrates the clear benefits of tool integration. Qwen2.5-7B and Kimi Audio’s accuracy rises substantially, underscoring the value of leveraging external context. As expected, latency increases due to the additional overhead of retrieving information from external tools (See T.~\ref{tab:latency} for detailed latency analysis).

Most notably, our post-training approach to build \emph{Model-Triggered Streaming RAG}, as described in S.~\ref{subsec:streaming_rag_post_training_method}, delivers significant advancements in both accuracy and efficiency. Qwen2.5-7B and OpusLM achieve significant accuracy improvements across both benchmarks. While the absolute accuracy scores may appear low, they are consistent with the evaluation results observed in the CRAG benchmark. Notably, Qwen2.5-7B with \emph{Model-Triggered Streaming RAG} achieves accuracy comparable to the open book performance of similarly sized LLMs reported in the original CRAG paper (i.e., 34.2 for Qwen2.5-7B vs. 32.1 for LLAMA-3 8B Instruct in \citep{yang2024crag}). 
Importantly, although post-training is performed exclusively on the synthetic dataset, we observe consistent and even greater improvements on the human-spoken benchmark, demonstrating the robustness and strong generalization capabilities of our method.  This setting also introduces substantial latency savings compared to the Open Book configuration, with Qwen2.5-7B and OpusLM achieving 20.7\% and 14.8\% reductions in first-token latency on the synthetic benchmark, and even greater savings on the human benchmark
\footnote{Our latency calculations on synthetic audio exclude end-point detection latency, which is required in all production systems. Since our streaming RAG approach enables processing without waiting for end-point detection, including this factor would further amplify the observed latency savings, as observed in higher latency savings for human-spoken audio, where endpoint detection errors often introduce trailing silence.}. These results demonstrate that our streaming RAG approach not only advances the accuracy of speech-in speech-out systems, but also optimizes response efficiency by enabling earlier and more effective prediction of tool queries. 

\subsection{Analysis: Modality Gap Between Speech and Text Output}
\label{subsec:results_rag_text}
\begin{table}[t]
\centering
\caption{First Token Latency breakdown, showing median (P50) and 90th percentile (P90) timings, for the Qwen2.5-7B on AudioCRAG-Synthetic.}
\label{tab:latency}
\resizebox{\linewidth}{!}{
\begin{tabular}{|l|l|c|cc|c|c|}
\hline
\multirow{3}{*}{Model} & \multirow{3}{*}{Setting} & \multirow{3}{*}{P} & \multicolumn{4}{c|}{Latency (sec)}\\
\cline{4-7}
 &  &  & \multicolumn{2}{c|}{Tool Use Latency}  & \multirow{2}{*}{Response Gen}  & \multirow{2}{*}{Total} \\
\cline{4-5}
 & &  & Query Gen & Tool Results Gen &  &   \\
\hline
\multirow{4}{*}{Qwen2.5-7B} & Open Book & P50 & 0.59 & 2.78 & \hphantom{0}2.52 &  5.90 \\
 &  (S.~\ref{subsec:tool_integrate_method}) & P90 & 0.85 & 4.90 & \hphantom{0}3.25 & 9.00 \\
\cline{2-7}
& Streaming RAG & P50 & 0.59 & 2.20 & \hphantom{0}2.52 &  5.32 \\
& (S.~\ref{subsec:streaming_rag_post_training_method}) & P90 & 0.85 & 4.37 & \hphantom{0}3.25 &  8.47\\
\hline
\end{tabular}
}
\end{table}
Table~\ref{table:speech_to_text_performance} provides a comparative evaluation of the three SOTA models in generating either text or speech outputs from speech inputs, both with and without the integration of external tool results. In the absence of tool results (Ref length = 0), all models achieve higher accuracy when generating text outputs compared to speech outputs. The incorporation of tool results generally leads to improved text generation performance, with Kimi Audio achieving the highest accuracy. In contrast, the accuracy for speech output remains consistently lower across all models and conditions. The self-cascade approach, in which the model first generates an intermediate text response before producing the final speech output, provides moderate improvements in speech output accuracy for both Qwen2.5-7B and OpusLM.
However, it still underperforms compared to the text-out baseline, primarily due to errors in accurately generating answers involving uncommon entity nouns. 
Overall, these findings underscore a persistent challenge in direct speech generation, particularly when tool results are integrated, as this appears to negatively impact the quality of generated speech responses.
Our \emph{Model-Triggered Streaming RAG} demonstrates clear advantages: it maintains comparable performance for text output and delivers substantial improvements for speech output, even outperforming the self-cascade approach. These results underscore the effectiveness of post-training in overcoming challenges in direct speech generation in tool-augmented scenarios.

\subsection{Analysis of Latency Bottlenecks in Tool-Integrated Speech Dialogue}

Table~\ref{tab:latency} provides a comprehensive breakdown of latency measurements for the Qwen2.5-7B speech-to-speech model, evaluated in both the Open Book setting and our proposed \emph{Model-Triggered Streaming RAG} setting on AudioCRAG-Synthetic. We report both median (P50) and 90th percentile (P90) values for each stage of the processing pipeline. The latency is decomposed into three main components: tool query generation, tool result generation, and speech response synthesis, with measurements provided for both the first token outputs (We also provide latency measurements for last token output in T.~\ref{tab:last_token_latency}). For both tool query and tool result generation, it is assumed that all tools are accessed in parallel; thus, the reported latency corresponds to the maximum query or result generation time among all tools. The majority of this latency arises from leveraging external web pages, which introduces significant delays, most notably, increasing the first token latency by 2.3x in Open Book Setting. 
Notably, our \emph{Model-Triggered Streaming RAG} setting enables early generation of tool results, successfully reducing P50 first token latency by 9.8\% and tool use latency by 20.7\%.

\subsection{Ablation Study}
\begin{figure}
\centering
\includegraphics[width=0.9\linewidth]{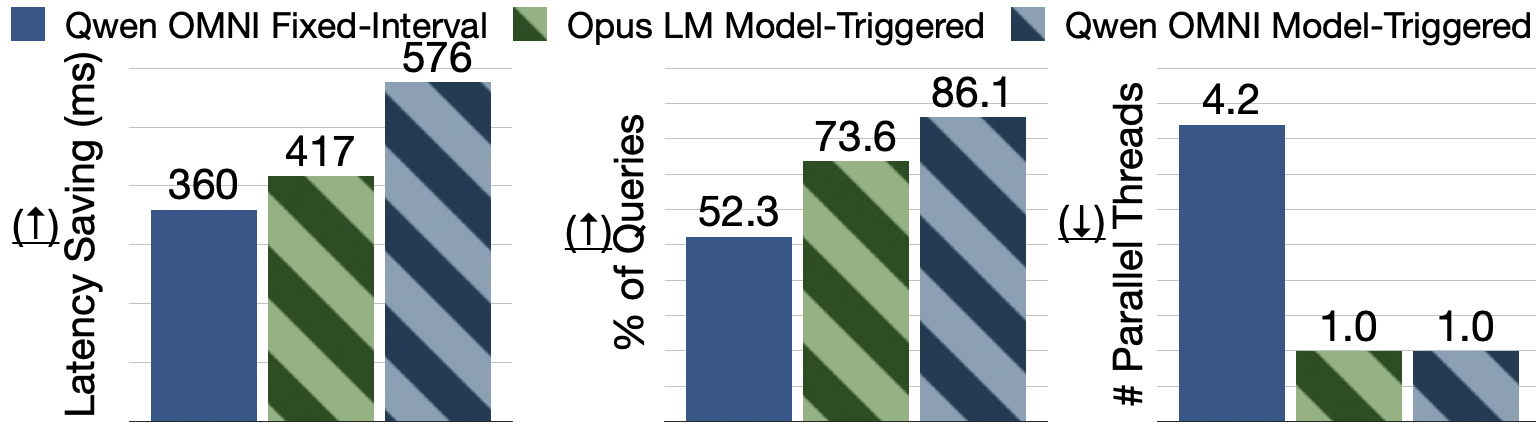}
\caption{Average latency savings (Mean) by streaming RAG approaches (S.~\ref{sec:stream_rag}.)}
\label{fig:streaming_rag_post_trainsavings}
\end{figure}
\textbf{Streaming RAG approaches}: Figure~\ref{fig:streaming_rag_post_trainsavings} highlights the substantial latency savings enabled by the \emph{Model-Triggered Streaming RAG} (S.~\ref{subsec:streaming_rag_post_training_method}), compared to the \emph{Fixed-Interval Streaming RAG} (S.~\ref{subsec:stream_rag_infer_method}). Three key metrics are evaluated: overall latency savings, the percentage of queries benefiting from reduced latency, and the number of parallel threads required.
Even without any post-training, the \emph{Fixed-Interval Streaming RAG} approach already reduces tool usage latency (3.37s in T.~\ref{tab:latency}) by 10.7\% for Open Book Qwen-OMNI, demonstrating its flexibility and plug-and-play compatibility with any existing speech-in speech-out model.
The \emph{Model-Triggered Streaming RAG} method, utilizing Qwen-OMNI, consistently delivers superior performance. It achieves greater average latency reductions and benefits a higher proportion of queries with improved response times. Notably, \emph{Model-Triggered Streaming RAG} require only a single parallel thread, representing a significant advancement in resource efficiency compared to the \emph{Fixed-Interval Streaming RAG} approach, which demands multiple parallel threads.
These findings highlight the effectiveness of \emph{Model-Triggered Streaming RAG} in not only minimizing latency, validating our intuition from Section~\ref{subsec:streaming_rag_post_training_method}, but also in optimizing system efficiency and resource utilization.

\begin{table}[t]
\centering
\caption{Comparison of speech-to-speech model under different post-training conditions}
\label{tab:post_train_ablation_performance}
\begin{tabular}{lllll}
\toprule
Post-Training & Ref length & Post Train Data & Model & Acc.  \\
\midrule
Tool Integration (S.~\ref{subsec:baseline_models}) & 15K & 16K  & OpusLM & 22.4 \\
Streaming RAG (S.~\ref{subsec:streaming_rag_post_training_method}) & 15K & 16K  & OpusLM  & 23.6 \\
Tool Integration (S.~\ref{subsec:baseline_models}) & 23K & 16K  & Qwen2.5-7B & 34.9 \\
Streaming RAG (S.~\ref{subsec:streaming_rag_post_training_method}) & 23K & 16K & Qwen2.5-7B  & 34.2 \\
\bottomrule
\end{tabular}
\end{table}
\textbf{Post-training Strategies}: Table \ref{tab:post_train_ablation_performance} presents performance comparison under different post-training conditions. Incorporating streaming tool query generation during post-training (S.~\ref{subsec:streaming_rag_post_training_method}) results in comparable performance for both models. These results suggest that \emph{Model-Triggered Streaming RAG} can be integrated into post-training process without negatively impacting model performance.

\begin{table}[t]
\centering
\caption{Ablation Results for Negative Sampling Strategy in \emph{Model-Triggered Streaming RAG} (S.~\ref{subsec:streaming_rag_post_training_method}).}
\begin{tabular}{lcccccc}
\toprule
 Scenario & Ref Length & Post Train Data & Output & Model & Acc. \\
\midrule
Open Book & 15K & 0    & Text & Qwen2.5-7B & 39.6 \\ \midrule
Post-train (S.~\ref{subsec:streaming_rag_post_training_method})  & 15K & 16K & Text & Qwen2.5-7B & 39.8 \\
\hphantom{0}- Negative sampling & 15K & 16K & Text & Qwen2.5-7B & 36.5 \\
\bottomrule
\end{tabular}
\label{tab:streaming_rag_ablatiob_results}
\end{table}
\textbf{Ablation Results for Negative Sampling Strategy}: Table~\ref{tab:streaming_rag_ablatiob_results} presents the ablation results evaluating the impact of deliberately injecting negative samples during post-training (Eq.~\ref{eq:stream_RAG_neg_sample}). The findings highlight that, without negative sampling, streaming tool query generation can reduce final accuracy in text output settings, primarily due to errors in final query generation, as detailed in S.~\ref{subsec:streaming_rag_post_training_method}. In contrast, our negative sampling approach significantly enhances the model’s robustness, enabling it to recover from intermediate prediction errors. This leads to consistently high accuracy while also achieving notable latency reductions (Tab.~\ref{tab:latency}).

\section{Conclusion and Discussion}
In this work, we introduced the first comprehensive approach for integrating external tool usage directly into E2E speech-in, speech-out dialogue systems. By incorporating \emph{Model-Triggered Streaming RAG} pipeline, we further enhanced the model’s ability to leverage retrieved information and autonomously decide when to trigger new tool queries, resulting in improved accuracy and responsiveness.
Empirical evaluation on the newly introduced AudioCRAG benchmark demonstrated that tool integration can more than double factual question answering accuracy compared to closed-book models. Additionally, our streaming RAG approach achieved a 20\% reduction in tool usage latency, thereby preserving natural conversational flow. 
Overall, our contributions advance the state of the art in spoken dialogue systems by enabling accurate, real-time, and tool-augmented voice assistants. In addition, we are committed to open science and will release our training code and Audio CRAG Human benchmark to support future research and development in this area.

\clearpage
\newpage
\bibliographystyle{assets/plainnat}
\bibliography{paper}

\clearpage
\newpage
\beginappendix

\section{Benchmarking Text Dialogue Systems for Tool Usage}
\label{subsec:related_text_dialogue_tool}
Recent advances in benchmarking text-based dialogue systems for tool usage \citep{text_rag_bench5,text_rag_bench1,text_rag_bench2,text_rag_bench3,text_rag_bench4,vu2023freshllmsrefreshinglargelanguage,xiong-etal-2024-interactive,peng2024graphretrievalaugmentedgenerationsurvey,su2025temporalknowledgegraphquestion,ni2025trustworthyretrievalaugmentedgeneration} have primarily focused on evaluating factual question answering and task completion within simulated environments. The CRAG benchmark \citep{yang2024crag} is a leading example, featuring 4,409 question-answer pairs and providing mock APIs for both web and knowledge graph (KG) search. CRAG supports a range of KG and web retrieval tasks, and highlights key challenges such as hallucinations in retrieval-augmented generation (RAG) and the importance of leveraging KGs and search ranking to improve factual accuracy. Evaluation is conducted automatically using two LLM judges. SimpleQA \citep{simpleqa} is another widely adopted benchmark, designed to assess language models on short, fact-seeking questions. With 4,326 adversarially collected questions spanning diverse topics and a straightforward grading scheme based on single, indisputable answers, SimpleQA provides a robust testbed for factual accuracy. Moving beyond question answering, WebArena \citep{webarena} offers a simulated environment for evaluating dialogue agents on web-based tasks using fully functional websites, enabling assessment of more complex, action-oriented behaviors. While these benchmarks have significantly advanced the evaluation of tool-augmented dialogue systems, they remain largely limited to text-based interactions and do not fully address the unique challenges presented by speech-in speech-out systems.
\section{Multimodal Benchmarks for Tool Usage}
\label{subsec:related_multimodal_tool}
Recent benchmarks~\cite{mei2025surveymultimodalretrievalaugmentedgeneration,yu2025visragvisionbasedretrievalaugmentedgeneration,luo2024videoragvisuallyalignedretrievalaugmentedlong} have extended tool-augmented dialogue evaluation to multimodal and longer-context scenarios. The m\&m’s benchmark \citep{ma2024mms} evaluates LLMs on multi-step, multi-modal tasks using a diverse set of 33 tools, including public APIs and multimodal models such as off-the-shelf automatic speech recognition (ASR) models, highlighting the potential for developing agents that leverage audio-based tools.
CRAG\_MM \citep{cragmm2025} builds on the original CRAG benchmark by introducing visual question answering (QA) tasks that combine images and text-based queries, utilizing mock APIs for both image descriptions and web search. 
For video understanding and long-context reasoning, the Video Web Arena \citep{videowebarena} benchmark evaluates multimodal agents on tasks involving 2,021 manually crafted tutorial videos. 
While these benchmarks advance the field by incorporating multimodal tools, they still do not evaluate systems in speech-in speech-out scenarios.

\section{Audio CRAG Benchmark}
\label{subsec:audio_crag_appendix}
We begin with the CRAG dataset \citep{yang2024crag}, which contains 2,706 text queries. Since these queries are not directly suitable for speech-based evaluation, we first identify those requiring adaptation before TTS conversion. Through careful manual inspection, we determine that queries containing dates or brackets benefit from rewriting to ensure naturalness and clarity in spoken form. In total, we identify 569 such queries and rewrite them using a large language model (LLAMA-4 Maverick). The resulting 569 rewritten queries, combined with the remaining original queries, form the basis for our spoken version of CRAG, which we term \emph{AudioCRAG}. We follow the CRAG setup to incorporate web and KG-based tools and adopt its robust evaluation setup, as detailed in S.~\ref{subsec:tool_usage_setup} and~\ref{subsec:evaluation_setting}. 

\textbf{Audio CRAG Synthetic}: 
To generate spoken queries, we process all 2,706 queries through our in-house TTS system. We then apply a rigorous filtering procedure to remove queries with intelligibility issues, specifically excluding any utterances for which Whisper~\cite{whisper} hypotheses exhibit a non-zero word error rate. We also remove utterances with suboptimal audio quality, as determined by UTMOS~\citep{saeki22c_interspeech} scores below 3.5. This results in a high-quality set of 1,862 spoken queries, which we refer to as the \emph{AudioCRAG-Synthetic} benchmark.

\textbf{Audio CRAG Human}: 
To further enhance the realism and diversity of our evaluation, we introduce the AudioCRAG-Human benchmark, which consists of 618 human-recorded spoken queries. These queries are recorded by a diverse pool of participants to capture natural variations in speech, accent, and prosody. The inclusion of human-recorded audio enables a more comprehensive assessment of speech-in speech-out systems under real-world conditions, providing valuable insights into model robustness and generalization beyond synthetic speech. This benchmark serves as a critical resource for evaluating the effectiveness of tool integration in conversational AI systems.

\section{Tool Usage Setup}
\label{subsec:tool_usage_setup}
To enable effective tool usage, we build upon the CRAG framework by integrating two complementary information sources: web search, which provides access to fresh and dynamic content, and a knowledge graph, which offers structured and reliable information. For web search, we aggregate all 100,000 documents from the CRAG corpus and employ a BGE-based re-ranker\footnote{\url{https://huggingface.co/BAAI/bge-large-en-v1.5}}~\cite{bge_embedding} to index and retrieve the top 50 most relevant documents for each query. These documents are then segmented into chunks and re-ranked using the same BGE model based on their similarity to the query, ensuring highly contextually relevant retrieval. Meanwhile, queries to the knowledge graph are performed via a simulated API, adhering to the methodology established in CRAG~\footnote{\url{https://github.com/facebookresearch/CRAG/tree/main/mock_api}}.
\begin{table}[h!]
\centering
\caption{Results on AudioCRAG-Synthetic for Qwen2.5-7B, OpusLM, and Kimi Audio comparing text vs. speech output across Open Book setting showing average rates of accurate, hallucinated, and missing responses, as well as overall
truthfulness scores for each system.}
\begin{tabular}{llllllll}
\toprule
Ref length & Output & Model      & Score  & Acc. & Halluc & Miss. \\
\midrule
0          & Text   & Qwen2.5-7B & -13.1  & 15.0 & 28.1   & 56.9  \\
0          & Speech & Qwen2.5-7B & -21.1  & 11.1 & 32.3   & 56.6  \\
0          & Text   & OpusLM     & -44.3  & 20.1 & 64.3   & 15.6  \\
0          & Speech & OpusLM     & -47.9  & 18.4 & 66.2   & 15.4  \\
0          & Text   & Kimi Audio & -38.5  & 24.2 & 62.7   & 13.1  \\
0          & Speech & Kimi Audio & -53.5  & 16.7 & 70.3   & 12.9  \\
\bottomrule
\end{tabular}
\label{table:model_performance}
\end{table}

\section{Evaluation Setting}
\label{subsec:evaluation_setting}
Similar to previous work \citep{yang2024crag}, we employ model-based automatic evaluation. We use a three-way scoring system, assigning scores of 1, -1, and 0 for accurate, incorrect, and missing answers, respectively. The evaluation is conducted using the Llama 4-maverick LLM evaluator. For speech outputs, we first transcribe the audio using Whisper~\citep{whisper} before passing the transcriptions to the LLM evaluator. In this study, our primary focus is on enhancing system accuracy; therefore, we report average accuracy values in Tables~\ref{table:main_model_performance} and~\ref{table:speech_to_text_performance}. For additional context, we also provide the average rates of accurate, hallucinated, and missing responses, as well as overall truthfulness scores for each system in the Open Book Setting (see Table~\ref{table:model_performance}). Notably, our results indicate that Qwen-OMNI was less likely to generate hallucinated responses compared to OpusLM and Kimi Audio.

\section{Experiment Setup of SOTA speech-in speech-out models}
\label{sec:app-1}

\textbf{Qwen-OMNI} \citep{Qwen2.5-Omni} is a end-to-end multimodal model that seamlessly integrates diverse input modalities—including text, images, audio, and video—and generates both text and natural speech responses in a real-time streaming fashion. It leverages an innovative Thinker-Talker architecture, where the Thinker module performs high-level reasoning to produce a text response, which is then used by the Talker module, conditioning on both the text and the Thinker's hidden representations, to generate streaming speech output.

\textbf{OpusLM} \citep{tian2025opuslm} is an open-source speech-in, speech-out model post-trained to directly answer complex semantic and factual questions from raw audio inputs, through Chain-of-Thought reasoning.

\textbf{Kimi Audio} \citep{ding2025kimi} is a  universal audio foundation model that unifies audio understanding, generation, and conversational abilities within a single framework. Pre-trained on over 13 million hours of diverse audio and text data, Kimi Audio achieves state-of-the-art performance across a wide range of audio benchmarks, including audio understanding and speech conversation tasks.

For tool-augmented scenarios, retrieval results are provided up to each model’s maximum token limit (``Ref length'' in Tables), maintaining a 2:1 ratio of web page to KG results.
Specifically, we observe that Kimi-Audio is currently optimized for handling tool result references up to a certain length. When this limit is exceeded, an error arises during the audio detokenization process, specifically within the rotary embedding mechanism, highlighting an architectural constraint in processing longer input sequences or larger reference contexts. Addressing this limitation presents a valuable opportunity for future model enhancements.

Our evaluation encompasses both speech-in-text-out and speech-in-speech-out scenarios. For the Fixed-Interval Streaming RAG setting (Section~\ref{subsec:stream_rag_infer_method}), intermediate tool queries are generated at consistent 1-second intervals. In the Model-Triggered Streaming RAG setting, the model dynamically determines the need for a tool call after processing each 500ms block. This approach allows us to utilize a smaller chunk size, as only a single tool call thread is required for Model-Triggered Streaming RAG, thereby enabling more efficient and responsive processing.

\section{Post Training Data Preparation}
\label{subsec:post_train_data_prep}
This subsection details the experimental setup for post-training the pretrained speech-in, speech-out model to significantly enhance its tool usage capabilities, as outlined in S.~\ref{subsec:streaming_rag_post_training_method}.
We leverage a random subset of 16,000 questions from the text-based factual question answering dataset TriviaQA~\citep{triviaqa}, which contains 97,000 questions and 662,659 associated web documents.
For \emph{Model-Triggered Streaming RAG}, we further compute word-level timestamps using a pre-trained ASR model, OWSM CTC v4 1B~\citep{owsm-v4}, enabling us to generate partial ASR transcripts $X^{\text{asr}}_b$ at 500 ms intervals. Note that if word occurs at boundary of block $b$, it is excluded from $X^{\text{asr}}_b$. For each partial transcript $X^{\text{asr}}_b$, we generate corresponding psuedo ground truth queries $\overline{Q^{T}_{b}}$ using LLAMA-4-Maverick to simulate incremental user input. 
To simulate realistic tool usage, we concatenate all documents and employ a web query reranker to retrieve the top 50 most relevant documents for each query. The text questions from this 16k subset are converted into discrete speech tokens using text-to-speech synthesis with the corresponding pretrained speech-in, speech-out model. Recognizing that TriviaQA answers are typically single named entities, we further transform the queries into a conversational style using LLAMA-4-Maverick, making them more suitable for dialogue-based evaluation.

\begin{table}[h!]
\centering
\caption{Example KG Queries, and Web Queries generated by Qwen-OMNI in Open Book setting}
\begin{tabular}{|p{5cm}|p{4cm}|p{4cm}|}
\hline
\textbf{ASR Transcript of Question $X^{\text{asr}}$} & \textbf{Web Query $\hat{Q}^{\text{web}}$} & \textbf{KG Query $\hat{Q}^{\text{KG}}$} \\
\hline
which of nolan greenwald's movies has achieved the highest level of box office success on a global scale? & Nolan Greenwald's highest-grossing movie & \{'domain': 'movie', 'movie\_name': "Nolan Greenwald's movies", 'movie\_aspect': 'revenue'\} \\
\hline
who has played drums for the red hot chili peppers? & Red Hot Chili Peppers drummers & \{'domain': 'music', 'artist\_name': 'Red Hot Chili Peppers', 'artist\_aspect': 'member'\} \\
\hline
what's the current stock price of tortoise midstream energy fund? & Tortoise Midstream Energy Fund stock price & \{'domain': 'finance', 'market\_identifier': 'Tortoise Midstream Energy Fund', 'metric': 'price', 'datetime': '02/28/2024'\} \\
\hline
what was the volume of trading in cabot corporation's stock on the most recent day that dividends were distributed? & CABOT Corp stock trading volume on dividend distribution date & \{'domain': 'finance', 'market\_identifier': 'Cabot Corporation', 'metric': 'dividend', 'datetime': '02/28/2024'\} \\
\hline
which movie won the academy award for best film in 2020? & 2020 Academy Award for Best Picture & \{'domain': 'movie', 'movie\_aspect': 'oscar\_awards', 'year': 2020\} \\
\hline
which teams have won against phoenix suns during 2022-12? & Teams that beat Phoenix Suns in December 2022 & \{'domain': 'sports', 'sport\_type': 'basketball', 'team': 'Phoenix Suns', 'datetime': '2022-12-15'\} \\
\hline
\end{tabular}
\label{tab:example_queries}
\end{table}

\begin{table}[h!]
\centering
\caption{Streaming Tool Queries generated by Qwen-OMNI after \emph{Model-Triggered Streaming RAG} post-training}
\begin{tabular}{|p{2cm}|p{4.5cm}|p{4.5cm}|p{4.5cm}|}
\hline
\textbf{ASR Transcript of Question $X^{\text{asr}}$} & \textbf{Partial ASR Transcript $X^{\text{asr}}_{b}$} & \textbf{Web Query $\hat{Q}^{\text{web}}_{b}$} & \textbf{KG Query $\hat{Q}^{\text{KG}}_{b}$} \\
\hline
\multirow{5}{2cm}{Who founded Rare Beauty in 2019?} 
    & Who founded & Who founded what & \{"domain": "other", "main\_entity": "Who"\} \\
    & Who founded rare & Red Bull founder & \{"domain": "other", "main\_entity": "Red"\} \\
    & Who founded rare beauty & Who founded Rare Beauty & \{"domain": "other", "main\_entity": "Rare Beauty"\} \\
    & Who founded rare beauty in & NO\_QUERY & NO\_QUERY \\
    & Who founded rare beauty in 2019? & NO\_QUERY & NO\_QUERY \\
\hline
\multirow{12}{2cm}{How many jump shots did Darius Miles score with in the game on 2000-11-08?}
    & how many & Number of queries on 03/27/2024 & \{"domain": "other", "main\_entity": "How many"\} \\
    & how many jump & Number of jump shots in basketball & \{"domain": "other", "main\_entity": "jump shot"\} \\
    & how many jump shots did & Derek Jeter jump shots count & \{"domain": "sports", "sport\_type": "other", "person": "Dairy", "datetime": "03/27/2024, 19:52:43 PT"\} \\
    & how many jump shots did Darius miles & Darius jump shots count & \{"domain": "sports", "sport\_type": "other", "person": "Darius", "datetime": "03/27/2024, 19:52:43 PT"\} \\
    & how many jump shots did Darius miles score & Darius Miles jump shots count & \{"domain": "sports", "sport\_type": "other", "person": "Darius Miles", "datetime": "03/27/2024, 19:52:43 PT"\} \\
    & how many jump shots did Darius miles score with in & Darius Miles jump shots scored & NO\_QUERY \\
    & how many jump shots did Darius miles score with in the game on & Darius Miles jump shots scored in game on 03/27/2024 & NO\_QUERY \\
    & how many jump shots did Darius miles score with in the game on November & Darius Miles jump shots scored in game on November 2024 & \{"domain": "sports", "sport\_type": "other", "person": "Darius Miles", "datetime": "November"\} \\
    & how many jump shots did Darius miles score with in the game on November 8 & Darius Miles jump shots scored on November 8 & \{"domain": "sports", "sport\_type": "other", "person": "Darius Miles", "datetime": "November 8"\} \\
    & how many jump shots did Darius miles score with in the game on November 8 & NO\_QUERY & NO\_QUERY \\
    & how many jump shots did Darius miles score with in the game on November 8, 2000 & Darius Miles jump shots scored on November 8, 2000 & \{"domain": "sports", "sport\_type": "other", "person": "Darius Miles", "datetime": "November 8, 2000"\} \\
\hline
\end{tabular}
\label{tab:example_stream_queries}
\end{table}

\begin{table}[t]
\centering
\caption{Last-token Latency breakdown, showing median (P50) and 90th percentile (P90) timings, for the Qwen2.5-7B in Open Book Setting on AudioCRAG-Synthetic (First Token Latency=5.9 sec in T.~\ref{table:main_model_performance}).}
\label{tab:last_token_latency}
\begin{tabular}{|l|l|c|cc|c|c|}
\hline
\multirow{3}{*}{Model} & \multirow{3}{*}{Token} & \multirow{3}{*}{P} & \multicolumn{4}{c|}{Latency (sec)}\\
\cline{4-7}
 &  &  & \multicolumn{2}{c|}{Tool Latency}  & \multirow{2}{*}{Response Gen}  & \multirow{2}{*}{Total} \\
\cline{4-5}
 & &  & Query Gen & Tool Results Gen &  &   \\
\hline
\multirow{2}{*}{Qwen2.5-7B} & \multirow{2}{*}{Last Token} & P50& 0.59 & 2.78 & 16.70 & 20.07  \\
 & & P90  & 0.85 & 4.90 & 42.41 & 48.16\\
\hline
\end{tabular}
\end{table}

\section{Prompts used for Factual QA}
\label{appendix:prompt}

\subsection{Prompt in Closed Book Setting.}
\label{appendix:prompt_llm}
PROMPT = """
You are given an Audio Question and the time when it was asked in the Pacific Time Zone (PT), referred to as "Query Time". The query time is formatted as "mm/dd/yyyy, hh:mm:ss PT". Your task is to answer the question in as few words as possible.\\
Please follow these guidelines when formulating your answer:\\
1. If the question contains a false premise or assumption, answer “invalid question”.\\
2. If you are uncertain or don’t know the answer, respond with “I don’t know”.\\
\#\#\# Question\\
\{query\}\\
\#\#\# Query Time\\
\{query\_time\}\\
\#\#\# Answer\\
"""

\subsection{Prompt in Open Book / Streaming RAG Setting.}
PROMPT = """
You are given an Audio Question, References and the time when it was asked in the Pacific Time Zone (PT), referred to as "Query Time". The query time is formatted as "mm/dd/yyyy, hh:mm:ss PT". The references may or may not help answer the question. Your task is to answer the question in as few words as possible.\\
Please follow these guidelines when formulating your answer:\\
1. If the question contains a false premise or assumption, answer “invalid question”.\\
2. If you are uncertain or don’t know the answer, respond with “I don’t know”.\\
\#\#\# Question\\
\{query\}\\
\#\#\# Query Time\\
\{query\_time\}\\
\#\#\# References\\
\# web\\
\{web\_results\}\\
\# knowledge graph\\
\{kg\_response\}\\
\#\#\# Answer\\
"""

\subsection{KG Query extraction in Open Book Setting.}
PROMPT = """
You are an agent that only outputs JSON. You are given a Query and Query Time. Do the following:\\\\
1) Determine the domain the query is about. The domain should be one of the following: "finance", "sports", "music", "movie", "encyclopedia". If none of the domains apply, use "other". Use "domain" as the key in the result json.\\\\
2) Extract structured information from the query. Include different keys into the result json depending on the domains, and put them DIRECTLY in the result json. Here are the rules:\\\\
For `encyclopedia' and `other' queries, these are possible keys:\\
-  `main\_entity': extract the main entity of the query.\\\\
For `finance' queries, these are possible keys:\\
- `market\_identifier': stock identifiers including individual company names, stock symbols.\\
- `metric': financial metrics that the query is asking about. This must be one of the following: `price', `dividend', `P/E ratio', `EPS', `marketCap', and `other'.\\
- `datetime': time frame that the query asks about. When datetime is not explicitly mentioned, use `Query Time' as default.\\\\
For `movie' queries, these are possible keys:\\
- `movie\_name': name of the movie\\
- `movie\_aspect': if the query is about a movie, which movie aspect the query asks. This must be one of the following: `budget', `genres', `original\_language', `original\_title', `release\_date', `revenue', `title', `cast', `crew', `rating', `length'.\\
- `person': person name related to moves\\
- `person\_aspect': if the query is about a person, which person aspect the query asks. This must be one of the following: `acted\_movies', `directed\_movies', `oscar\_awards', `birthday'.\\
- `year': if the query is about movies released in a specific year, extract the year\\\\
For `music' queries, these are possible keys:\\
- `artist\_name': name of the artist\\
- `artist\_aspect': if the query is about an artist, extract the aspect of the artist. This must be one of the following: `member', `birth place', `birth date', `lifespan', `artist work', `grammy award count', `grammy award date'.\\
- `song\_name': name of the song\\
- `song\_aspect': if the query is about a song, extract the aspect of the song. This must be one of the following: `author', `grammy award count', `release country', `release date'.\\\\
For `sports' queries, these are possible keys:\\
- `sport\_type': one of `basketball`, `soccer`, `other`\\
- `tournament': NBA, World Cup, Olympic.\\
- `team': teams that users are interested in.\\
- `datetime': time frame that the user is interested in. When datetime is not explicitly mentioned, use `Query Time' as default.\\\\
Return the results in a FLAT json.\\\\
*NEVER include ANY EXPLANATION or NOTE in the output, ONLY OUTPUT JSON!!!*\\
"""

\subsection{KG Query extraction in Streaming RAG Setting.}
PROMPT = """
You are an agent that only outputs JSON. You are given an Audio
Query, Previously generated JSON result ('Previous Result') and Query Time. Do the following:

1) Determine the domain the query is about. The domain should be one of the following:
\"finance\", \"sports\", \"music\", \"movie\", \"encyclopedia\". If none of the domains apply, use \"other\". Use
\"domain\" as the key in the result json.

2) Extract structured information from the query. Include different keys into the result json
depending on the domains, and put them DIRECTLY in the result json. Here are the rules:

For ‘encyclopedia’ and ‘other’ queries, these are possible keys:
- ‘main\_entity’: extract the main entity of the query.

For ‘finance’ queries, these are possible keys:
- ‘market\_identifier’: stock identifiers including individual company names, stock symbols.
- ‘metric’: financial metrics that the query is asking about. This must be one of the following: ‘price’,
‘dividend’, ‘P/E ratio’, ‘EPS’, ‘marketCap’, and ‘other’.
- ‘datetime’: time frame that the query asks about. When datetime is not explicitly mentioned, use
‘Query Time’ as default.

For ‘movie’ queries, these are possible keys:
- ‘movie\_name’: name of the movie
- ‘movie\_aspect’: if the query is about a movie, which movie aspect the query asks. This must be one
of the following: ‘budget’, ‘genres’, ‘original\_language’, ‘original\_title’, ‘release\_date’, ‘revenue’,
‘title’, ‘cast’, ‘crew’, ‘rating’, ‘length’.
- ‘person’: person name related to moves
- ‘person\_aspect’: if the query is about a person, which person aspect the query asks. This must be
one of the following: ‘acted\_movies’, ‘directed\_movies’, ‘oscar\_awards’, ‘birthday’.
- ‘year’: if the query is about movies released in a specific year, extract the year

For ‘music’ queries, these are possible keys:
- ‘artist\_name’: name of the artist
- ‘artist\_aspect’: if the query is about an artist, extract the aspect of the artist. This must be one of the
following: ‘member’, ‘birth place’, ‘birth date’, ‘lifespan’, ‘artist work’, ‘grammy award count’,
‘grammy award date’.
- ‘song\_name’: name of the song
- ‘song\_aspect’: if the query is about a song, extract the aspect of the song. This must be one of the
following: ‘author’, ‘grammy award count’, ‘release country’, ‘release date’.

For ‘sports’ queries, these are possible keys:
- ‘sport\_type’: one of ‘basketball‘, ‘soccer‘, ‘other‘
- ‘tournament’: NBA, World Cup, Olympic.
- ‘team’: teams that users are interested in.
- ‘datetime’: time frame that the user is interested in. When datetime is not explicitly mentioned, use
‘Query Time’ as default.
Return the results in a FLAT json.\\
*NEVER include ANY EXPLANATION or NOTE in the output, ONLY OUTPUT JSON!!!*

3) Compare your newly generated result to the 'Previous Result'. **If your new result would be exactly the same as the 'Previous Result', output only NO\_QUERY.**
Return the results in a FLAT json.

Previous Result: \\
\{prev\_kg\_query\}

"""

\subsection{Web Query extraction in Open Book Setting.}
PROMPT = """ You are given an Audio Query and Query Time. Your task is to generate a web query that can be used to retrieve relevant web pages. Rewrite the following query into a short and succinct form, focusing on the main topic or domain (e.g. finance, sports, music, movie, encyclopedia), key entities mentioned (e.g. people, organizations, locations), and specific aspects of those entities (e.g. performance metrics, relationships, events). Ensure the rewritten query is clear, concise, and easy to understand.
Note that simply outputting the original query is not acceptable. You must rephrase the query to make it more concise and focused on the key information that will help retrieve relevant web pages.

For 'finance' queries, focus on:
- Company names or stock symbols
- Financial metrics (e.g. price, dividend, P/E ratio, EPS, marketCap)
- Specific timeframes or events; if no timeframe is specified, use the Query Time as default

For 'sports' queries, focus on:
- Sports Type (eg. basketball, soccer)
- Teams, players
- Statistics or performance metrics (e.g. scores, wins, losses)
- Specific events or tournaments (eg. NBA, World Cup, Olympic)
- Time frame that the user is interested in; if no timeframe is specified, use the Query Time as default

For 'music' queries, focus on:
- Artist names or song titles
- Specific aspects of artist (eg. band name, birth place, birth date, lifespan, artist work, grammy award count, grammy award date)
- Specific aspects of song (eg. author, grammy award count, release country, release date)
- Music genres or categories
- Specific awards or recognition (e.g. Grammy Awards, Billboard)

For 'movie' queries, focus on:
- Movie titles or celebrity names
- Movie genres or other categories like budget, language, release\_date, revenue, cast, crew, rating, length
- Specific aspects of celebrity like acted\_movies, directed\_movies, oscar\_awards, birthday
- Specific awards or recognition (e.g. Oscars)
For 'other' queries, focus on:
- Main entity or topic
- Specific aspects or attributes of the entity

When rewriting the query, ensure that it captures all important information from the original question that could impact the retrieval results. Do not omit any crucial details, such as specific dates, locations, or relationships between entities. Also, do not invent any new details on your own. If necessary, use the Query Time to provide context for the query. The goal is to create a concise and accurate query that effectively conveys the user's intent and retrieves relevant information.
*NEVER include ANY EXPLANATION or NOTE in the output, ONLY OUTPUT THE REWRITTEN QUERY!!!*
"""

\subsection{Web Query extraction in Streaming RAG Setting.}
PROMPT = """You are given an Audio Query, previously generated Web query ('Previous Result') and Query Time. 

Your task is to generate a web query that can be used to retrieve relevant web pages. Rewrite the following query into a short and succinct form, focusing on the main topic or domain (e.g. finance, sports, music, movie, encyclopedia), key entities mentioned (e.g. people, organizations, locations), and specific aspects of those entities (e.g. performance metrics, relationships, events). Ensure the rewritten query is clear, concise, and easy to understand.

Note that simply outputting the original query is not acceptable. You must rephrase the query to make it more concise and focused on the key information that will help retrieve relevant web pages.

For 'finance' queries, focus on:
- Company names or stock symbols
- Financial metrics (e.g. price, dividend, P/E ratio, EPS, marketCap)
- Specific timeframes or events; if no timeframe is specified, use the Query Time as default

For 'sports' queries, focus on:
- Sports Type (eg. basketball, soccer)
- Teams, players
- Statistics or performance metrics (e.g. scores, wins, losses)
- Specific events or tournaments (eg. NBA, World Cup, Olympic)
- Time frame that the user is interested in; if no timeframe is specified, use the Query Time as default

For 'music' queries, focus on:
- Artist names or song titles
- Specific aspects of artist (eg. band name, birth place, birth date, lifespan, artist work, grammy award count, grammy award date)
- Specific aspects of song (eg. author, grammy award count, release country, release date)
- Music genres or categories
- Specific awards or recognition (e.g. Grammy Awards, Billboard)

For 'movie' queries, focus on:
- Movie titles or celebrity names
- Movie genres or other categories like budget, language, release\_date, revenue, cast, crew, rating, length
- Specific aspects of celebrity like acted\_movies, directed\_movies, oscar\_awards, birthday
- Specific awards or recognition (e.g. Oscars)

For 'other' queries, focus on:
- Main entity or topic
- Specific aspects or attributes of the entity

When rewriting the query, ensure that it captures all important information from the original question that could impact the retrieval results. Do not omit any crucial details, such as specific dates, locations, or relationships between entities. Also, do not invent any new details on your own. If necessary, use the Query Time to provide context for the query. The goal is to create a concise and accurate query that effectively conveys the user's intent and retrieves relevant information.
Now, compare the new web query to the previously generated web query  ('Previous Result').

If the new query is similar enough to the previous web query (i.e., it effectively conveys the same user intent and would retrieve similar relevant information), output only *NO\_QUERY*.

Previous Result: \\
\{prev\_web\_query\}\\
"""

\section{LLM as judge prompt}
\label{subsec:appendix_llm_judge_prompt}
PROMPT="""\\
Assume you are a human expert in grading predictions given by a model. You are given a question and a model prediction. Judge if the prediction matches the ground truth answer by following these steps:
1: Take it as granted that the Ground Truth is always correct.
2: If the Prediction indicates it is not sure about the answer, "score" should be "0"; otherwise, go the next step.
3: If the Prediction exactly matches the Ground Truth, "score" is 1.
4: If the Prediction does not exactly match the Ground Truth, go through the following steps and likely give a score as 0.
5: If the Ground Truth is a number, "score" is 1 if and only if the Prediction gives a number that almost exactly matches the ground truth.
6: If the Prediction is self-contradictory, "score" must be 0.
7: If the prediction is not answering the question, "score" must be 0.
8: If the prediction is a concise and correct summary of the ground truth, "score" is 1.
9: If ground truth contains a set of items, prediction must contain exactly same items for the score to be 1.
10: Otherwise, "score" is 0.

Output a JSON blob with an "explanation" field explaining your answer as short as possible and an "score" field with value 1 or 0.\\

You should make the judgment based on provided examples.
Examples:
Question: "which company has higher eps, btu or cma?"
Ground Truth: "cma"
Prediction: "it is not possible to determine which company has a higher eps."
Output: {"score": 0, "explanation": "The prediction is not sure about the answer."}

Question: "who is taller, a or b?"
Ground Truth: "a"
Prediction: "The answer is a. a is of height 1.75 m and b is of height 1.82 m. So b is taller."
Output: {"score": 0, "explanation": "The prediction is self-contradictory."}

Question: "who among patricia place and lana clarkson is younger?"
Ground Truth: "lana clarkson"
Prediction: "lana clarkson (born april 5, 1962) is younger than patricia clarkson (born december 29, 1959)."
Output: {"score": 1, "explanation": "The prediction gives the correct answer."}

Question: "did lakers win more games than warriors in 2024-4?"
Ground Truth: "yes"
Prediction: "yes, lakers won more games than warriors in 2023-2024 season."
Output: {"score": 0, "explanation": "The prediction is not answering the question."}

Question: "what is the height of the building where citigroup is headquartered?"
Ground Truth: "151 m"
Prediction: "the height is 915 feet (279 m)."
Output: {"score": 0, "explanation": "The prediction, 151 m, does not match the ground truth, 279 m."}

Question: "who were the last 3 prime ministers of uk?"
Ground Truth: "rishi sunak, liz truss, and boris johnson"
Prediction: "1. liz truss, 2. boris johnson, 3.theresa may"
Output: {"score": 0, "explanation": "The prediction does not contain item, rishi sunak, that is in the grount truth."}

Question: "who authored the taming of the shrew?"
Ground Truth: "william shakespeare"
Prediction: "william shakespeare"
Output: {"score": 1, "explanation": "The prediction exactly matches the ground truth."}

Question: "who authored The Taming of the Shrew?"
Ground Truth: "william shakespeare"
Prediction: "w shakespeare"
Output: {"score": 1, "explanation": "The prediction matches the ground truth as w is the abbreviation of william."}

Question: "what is the state bird of california?"
Ground Truth: "california quail"
Prediction: "california valley quail"
Output: {"score": 1, "explanation": "The prediction matches the ground truth as california quail is also known as california valley quail."}

Question: "what country is maharashtra metro rail corporation limited located in?"
Ground truth: "india"
Prediction: "indianapolis"
Output: {"score": 0, "explanation": "The prediction doesn't match the ground truth answers."}

Question: "what was wolves's record in their last game in eng-premier league?"
Ground Truth: "w"
Prediction: "wolves's record in their last game in the eng-premier League was a 2-1 win over fulham."
Output: {"score": 1, "explanation": "The prediction correctly answers the result is a win."}

Question: "how many seconds is 3 minutes 15 seconds?"
Ground Truth: "195 seconds"
Prediction: "3 minutes 15 seconds is 195 seconds."
Output: {"score": 1, "explanation": "The prediction matches ground truth as it correctly states the number of seconds asked in the question."}

Question: "on which days did am distribute dividends in the last year?"
Ground Truth: "2023-01-24, 2023-04-25, 2023-07-25, 2023-10-24"
Prediction: "the company distributed dividends on october 24, 2023 and january 24, 2023."
Output: {"score": 0, "explanation": "The prediction partially matches the ground truth, but it is incomplete."}

Question: "on which days did xxx distribute dividends in the last year?"
Ground Truth: "2023-01-13, 2023-03-25, 2023-11-21"
Prediction: "xxx distributed dividends on 1. 2023-01-13, 2. 2023-03-25, 3. 2023-10-21."
Output: {"score": 0, "explanation": "The prediction contains an item, 2023-10-21, which is not in ground truth."}

Question: "who are current band members of maroon 5?"
Ground Truth: "adam Levine, jesse carmichael, james valentine, matt flynn, pj morton, sam farrar"
Prediction: "they are jesse carmichael, matt flynn, adam levine, pj morton, sam farrar, and james valentine."
Output: {"score": 1, "explanation": "The prediction exactly matches the ground truth."}

Question: "which movies comprise the matrix franchise?"
Ground Truth: "the matrix, the matrix reloaded, the matrix revolutions, the matrix resurrections"
Prediction: "the matrix, the matrix reloaded, the matrix revolutions, the animatrix, and the matrix resurrections."
Output: {"score": 0, "explanation": "The prediction covers more items than what are given by the ground truth."}

Question: "how deep is the deepest lake of new york?"
Ground Truth: "618 ft"
Prediction: "the deepest lake in new york is seneca lake, with a depth of 618.23 feet."
Output: {"score": 1, "explanation": "The prediction exactly matches the number in ground truth after rounding."}

Question: "what is the closing price of meta yesterday?"
Ground Truth: "\$310.17"
Prediction: "310.2"
Output: {"score": 1, "explanation": "The prediction exactly matches the number in ground truth after rounding."}

Question: "what is the current market cap of appl?"
Ground Truth: "2.81 trillion"
Prediction: "2.667 trillion"
Output: {"score": 0, "explanation": "The prediction does not match the number in ground truth."}

Question: "what is the current pe ratio of appl?"
Ground Truth: "28.3"
Prediction: "the current pe ratio of apple is 26.66"
Output: {"score": 0, "explanation": "The prediction does not match the number in ground truth."}

Question: "how much is tesla's stock price down from its all-time high?"
Ground Truth: "\$221.83"
Prediction: "209.52"
Output: {"score": 0, "explanation": "The prediction does not match the number in ground truth."}

Question: "what is the length of amazon river?"
Ground Truth: "over 4000 miles"
Prediction: "the length of amazon river is 4,000 miles"
Output: {"score": 0, "explanation": "The prediction does not say Amazon River is longer than 4000 miles."}

Question: "how many copies x were sold?"
Ground Truth: "2 million."
Prediction: "it is over 2 million"
Output: {"score": 0, "explanation": "The prediction does not match the ground truth."}

Question: "what is the population of country x?"
Ground Truth: "3,576,873"
Prediction: "the population of country x is 3.3 million."
Output: {"score": 0, "explanation": "The prediction, 3.3 M, does not match the number, 3.6 M, in ground truth."}

Question: "what is the current market value of stock x?"
Ground Truth: "\$2,237,578,268"
Prediction: "\$2.16 billion."
Output: {"score": 0, "explanation": "The prediction, 2.16 B, does not match the number, 2.2 B, in ground truth."}

Question: "how old is michael jordan?"
Ground Truth: "61"
Prediction: "mj was born on February 17, 1963."
Output: {"score": 0, "explanation": "The prediction does not explicitly say the age."}

Question: "how long did amy teach as a teacher?"
Ground Truth: "24 years and 255 days"
Prediction: "25 years."
Output: {"score": 0, "explanation": "The prediction does not exactly match the ground truth."}

Question: "at what age did usher perform in the half time show of superbowl?"
Ground Truth: "45 years old"
Prediction: "usher has not performed at a super bowl halftime show."
Output: {"score": 0, "explanation": "The prediction does not match the ground truth."}

Question: "what year did olivia rodrigo win the grammys best new artist award?"
Ground Truth: "2022"
Prediction: "sure! here's an answer in three words or fewer: 2022 grammy awards."
Output: {"score": 1, "explanation": "The prediction answers the correct number of the year."}

Question: "how many times have zendaya and tom holland starred in a film together?"
Ground Truth: "3"
Prediction: "they have starred in 3 films together: spider-man: homecoming, spider-man: far from home, and spider-man: no way home."
Output: {"score": 1, "explanation": "The prediction gives the right number and it is reasonable."}

Question: "who is younger, timothee chalamet or tom holland?"
Ground Truth: "tom holland"
Prediction: "timothée chalamet is younger than tom holland."
Output: {"score": 0, "explanation": "The prediction does not match the ground truth."}

Question: "who had more number one hits on the us billboard, a or b?"
Ground Truth: "a had more number one hits on the us billboard than b, with 20 number one hits compared to b's 15."
Prediction: "a"
Output: {"score": 1, "explanation": "The prediction is a concise and correct summary of the ground truth."}

Question: "what is xxx's birthdate?"
Ground Truth: "1996-01-01."
Prediction: "02/01/1996"
Output: {"score": 0, "explanation": "The prediction does not match the ground truth."}

Question: "what was the worldwide box office haul for movie x?"
Ground Truth: "101756123."
Prediction: "102 million"
Output: {"score": 1, "explanation": "The prediction exactly matches the number in ground truth after rounding."}

Question: "how much has spotify's user base increased by since 2020 in na?"
Ground Truth: "spotify's user base increased by 34 million since 2020."
Prediction: "spotify's north american user base increased from 36 million in 2020 to 85 million by 2021"
Output: {"score": 0, "explanation": "The prediction is not answering the question as it only gives the increase from 2020 to 2021."}\\
"""
\begin{table}[h!]
\centering
\caption{Post-training Parameters of OpusLM}
\begin{tabular}{|p{5cm}|p{7cm}|}
\hline
\textbf{Parameter} & \textbf{Value} \\
\hline
train\_micro\_batch\_size\_per\_gpu & 1 \\
\hline
gradient\_accumulation\_steps & 2 \\
\hline
epochs & 2 \\
\hline
gradient\_clipping & 1.0 \\
\hline
bf16 enabled & true \\
\hline
optimizer type & Adam \\
\hline
optimizer lr & 0.00001 \\
\hline
optimizer betas & [0.9, 0.95] \\
\hline
optimizer eps & 1e-8 \\
\hline
optimizer weight\_decay & 3e-7 \\
\hline
optimizer adam\_w\_mode & true \\
\hline
scheduler type & WarmupDecayLR \\
\hline
scheduler warmup\_type & linear \\
\hline
scheduler total\_num\_steps & 21534 \\
\hline
scheduler warmup\_num\_steps & 1077 \\
\hline
scheduler warmup\_min\_lr & 0 \\
\hline
scheduler warmup\_max\_lr & 0.00001 \\
\hline
\end{tabular}
\label{tab:opuslm_posttrain_params}
\end{table}

\begin{table}[h!]
\centering
\caption{Post-training Parameters of Qwen-OMNI Thinker}
\begin{tabular}{|p{5cm}|p{7cm}|}
\hline
\textbf{Parameter} & \textbf{Value} \\
\hline
bf16 & True \\
\hline
gradient\_accumulation\_steps & 4 \\
\hline
epochs & 1 \\
\hline
gradient\_clipping & 1.0 \\
\hline
learning\_rate & 7e-6 \\
\hline
lr\_scheduler\_type & cosine \\
\hline
warmup\_ratio & 0.05 \\
\hline
per\_device\_train\_batch\_size & 1 \\
\hline
weight\_decay & 0.01 \\
\hline
\end{tabular}
\label{tab:qwen_omni_training_args}
\end{table}

\begin{table}[h!]
\centering
\caption{Post-training Parameters of Qwen-OMNI Talker}
\begin{tabular}{|p{5cm}|p{7cm}|}
\hline
\textbf{Parameter} & \textbf{Value} \\
\hline
bf16 & True \\
\hline
gradient\_accumulation\_steps & 4 \\
\hline
gradient\_clipping & 1.0 \\
\hline
epochs & 2 \\
\hline
learning\_rate & 5e-5\\
\hline
per\_device\_train\_batch\_size & 1 \\
\hline
lr\_scheduler\_type & linear \\
\hline
warmup\_ratio & 0.0 \\
\hline
weight\_decay & 0.01 \\
\hline
\end{tabular}
\label{tab:qwen_omni_talker_training_args}
\end{table}
\end{document}